\pdfoutput=1

\documentclass[11pt]{article}

\usepackage[]{EMNLP2022}

\usepackage{times}
\usepackage{latexsym}
\usepackage{makecell}
\usepackage{multirow}
\usepackage{caption}
\usepackage{subcaption}
\usepackage{overpic}
\usepackage{float}
\usepackage[T1]{fontenc}

\usepackage[utf8]{inputenc}

\usepackage{microtype}

\usepackage{inconsolata}

\title{An Empirical  Study of Automatic Post-Editing}

\author{Xu Zhang \\ Peking University \\ zhangxu@pku.edu.cn
        \And 
        Xiaojun Wan \\ Peking University \\ wanxiaojun@pku.edu.cn}

\begin{document}
\maketitle
\begin{abstract}
Automatic post-editing (APE) aims to reduce manual post-editing efforts by automatically correcting errors in machine-translated output. 
Due to the limited amount of human-annotated training data, data scarcity is one of the main challenges faced by all APE systems. 
To alleviate the lack of genuine training data, most of the current APE systems employ data augmentation methods to generate large-scale artificial corpora. 
In view of the importance of data augmentation in APE, we separately study the impact of the construction method of artificial corpora and artificial data domain on the performance of APE models.
Moreover, the difficulty of APE varies between different machine translation (MT) systems.
We study the outputs of the state-of-art APE model on a difficult APE dataset to analyze the problems in existing APE systems.
Primarily, we find that 
1) Artificial corpora with high-quality source text and machine-translated text more effectively improve the performance of APE models; 
2) In-domain artificial training data can better improve the performance of APE models, while irrelevant out-of-domain data actually interfere with the model; 
3) Existing APE model struggles with cases containing long source text or high-quality machine-translated text; 
4) The state-of-art APE model works well on grammatical and semantic addition problems, but the output is prone to entity and semantic omission errors.  
\end{abstract}

\section{Introduction}

The goal of automatic post-editing (APE; \citealp{simard-etal-2007-rule}) is to correct errors in machine-translated outputs that cannot be processed in the decoding stage \citep{bojar-etal-2017-findings}, thereby reducing human post-editing effort. 
APE systems learn knowledge from human post-edits, and apply it to the output of machine translation systems to improve the quality of translation. 

Human-annotated APE datasets are composed of triplets including source text (\textit{src}), machine-translated text (\textit{mt}), and human post-edits (\textit{pe}). 
APE models take \textit{src} and \textit{mt} as input and generate post-edited text with \textit{pe} as the target. 
During human post-editing, \textit{pe} is generated with reference to \textit{mt}, which means that different machine-translated outputs correspond to different human post-edits. 
Parallel corpora, which comprise pairs of source text (\textit{src}) and reference text (\textit{ref}), were utilized in a number of works \citep{1803.07274,lee-etal-2021-adaptation,2204.03896}.
\citet{lee-etal-2021-adaptation,2204.03896} have presented the edit distance between \textit{mt} and its target (\textit{pe} or \textit{ref} ) in detail to prove that \textit{pe} is closer to \textit{mt}.

In recent years, several studies have come to the conclusion that existing neural APE systems do not perform as well on strong in-domain neural machine translation (NMT) systems as on statistical machine translation (SMT) systems \citep{chatterjee-etal-2018-findings, chatterjee-etal-2019-findings}. 
As opposed to this view, one previous study \citep{chollampatt-etal-2020-automatic} pointed out that existing APE systems would still be able to improve the translation quality of strong NMT systems, given sufficient in-domain human-annotated APE triplets. 
However, we suppose it impractical to build different human post-edited datasets for different domains on all language pairs. 
Therefore, constructing large-scale APE datasets by manual post-editing is not the final resort for APE.

However, due to the scarcity of human post-edits (\textit{pe}), the quantity of genuine APE triplets in most APE datsets is heavily insufficient to train a Transformer-based Seq2Seq APE model. 
A number of studies \citep{1605.04800, 1803.07274, 2204.03896} circumvent the data scarcity problem by adopting data augmentation methods to generate artificial training data. 
We believe that the construction of artificial training data has a great impact on the performance of APE systems and higher-quality synthetic data can help improve APE system performance. 
To study data augmentation on APE task, we vary construction method of the artificial training data and domain of our synthetic data, and compare the final performance on multiple APE datasets. Empirically, we find that: 1) Construction methods that generate high-quality \textit{src} and \textit{mt} work best on APE models. 2) In-domain artificial training data are more beneficial to APE systems, while irrelevant out-of-domain synthetic data will hurt the performance of APE models. 

In addition, considering the fact that existing APE systems perform poorly on the output of strong NMT systems, we bring in an APE dataset constructed on the basis of the state-of-art NMT system, MLQE-PE \citep{2010.04480}. 
We compare the output of the state-of-art APE model on MLQE-PE test set with those on test sets of other APE datasets. 
In general, we find that despite the APE model performs quite differently on these datasets, it always struggles when dealing with triplets containing long \textit{src} or high-quality \textit{mt}. 
Moreover, we conduct human evaluation to analyze error type in \textit{mt}, the APE output and \textit{pe}. 
The experimental results demonstrate that the APE model works well on correcting grammatical and semantic addition problems in \textit{mt}, but fails on semantic omission and entity errors. 
We believe that these phenomena indicate the existence of some common problems in existing APE systems which need to be solved in future works. 
The code and data will be released to the community. 

\section{Related Work}

In contrast to \citet{pal-etal-2016-neural} formalizing APE as a monolingual MT problem in the target language, \citet{vu-haffari-2018-automatic} has applied the Seq2Seq model to APE with conditioning on \textit{src}. 
Early-stage neural APE models implemented a bidirectional RNN architecture \citep{pal-etal-2016-neural} to improve the translation output of SMT systems. 
On the basis of this model framework, a number of studies have separately imported alignment information \citep{pal-etal-2017-neural}, self-attention module \citep{junczys-dowmunt-grundkiewicz-2017-exploration}, attention query mechanism \citep{grangier-auli-2018-quickedit} and some other methods to improve the performance of APE models. 

Data scarcity has always been a tough issue for APE task because of the high cost of manual construction. 
Researchers have built several public artificial corpora to handle data scarcity, which have been widely used. 
\citet{1605.04800} and \citet{1803.07274} proposed two public large-scale artificial APE datasets constructed by round-trip translation and direct translation respectively. 
Inspired by back translation \citep{sennrich-etal-2016-improving} from MT task, \citet{lee-etal-2021-adaptation} adopted forward generation and backward generation to optimize the eSCAPE \citep{1803.07274} corpus. 

Transformer-based \citep{1706.03762} architecture has shown superiority on APE task. 
Dual-Source Transformer \citep{1809.00188} implemented two encoders to extract semantic information from \textit{src} and \textit{mt} respectively. \citet{huang-etal-2019-learning-copy} added copy mechanism to the Transformer architecture to help retain correctly translated parts in \textit{mt}. 
\citet{1906.06253} proposed \textit{BERT Enc.+BERT Dec.} model, which innovatively applied BERT \citep{1810.04805} to decoder and added parameter sharing \citep{sachan-neubig-2018-parameter} between encoder and decoder.

Although these Transformer-based APE models work well in many cases, they underperform on the output of some strong NMT systems \citep{ive-etal-2020-post}. 
Recent WMT APE task has focused on NMT systems, and the performance of existing models is not as good. 

\section{Experimental Setup}

\begin{table*}
\centering 
\begin{tabular}{cccccc} 
\hline
Dataset & Domain & Translation System & Train. size & Dev. size & Test. size \\
\hline
WMT'18 SMT & IT & SMT & 23,000 & 1,000 & 2,000 \\
SubEdits & Subtitles & NMT & 141,413 & 10,000 & 10,000 \\
MLQE-PE & Wikipedia & NMT & 7,000 & 1,000 & 1,000 \\
eSCAPE$^*$ & Mixed & NMT & 7,258,533 & / & / \\
SubEscape$^*$ & Subtitles & NMT & 5,633,518 & 10,000 & / \\
\hline

\end{tabular}
 \\
\caption{\label{tab:APE_data}
Summary of APE datasets used in this study, datasets marked with * are artificial APE data
}
\end{table*}

Considering the fact that more en-de APE datasets are available, we mainly focus on APE from English to German in our experiments. 
Two representative Transformer-based APE models are used to investigate the effect of different artificial training data. 
Two widely-used artificial corpora are utilized in our study as well.

\subsection{Datasets}

APE datasets are dividied into SMT APE datasets and NMT APE datasets according to the type of the MT system that generates \textit{mt}. 
The datasets we use for testing in our experiments include WMT'18 SMT, SubEdits and MLQE-PE. 
Since triplets in MLQE-PE are collected from Wikipedia containing various domains, it is not convenient to study data domain on this dataset. 
Therefore, we conduct comparative experiments on WMT'18 SMT and SubEdits.
In addition, constructed with a strong in-domain NMT system, MLQE-PE is more difficult for APE models than the other two datasets.
We use MLQE-PE to analyze the problems of the APE model that achieves the best results on WMT'18 SMT and SubEdits.
We also make use of two artificial APE datasets, eSCAPE \citep{1803.07274} and SubEscape \citep{chollampatt-etal-2020-automatic}, to help build different synthetic training data. 
The summary of these APE datasets is shown in Table \ref{tab:APE_data}.

\paragraph{WMT'18 SMT}   
WMT'18 SMT \citep{chatterjee-etal-2018-findings} is the data from the WMT 2018 APE shared task(en-de SMT), which consists of 23,000 triplets
for training, 1,000 for validation, and 2,000 for
testing.

\paragraph{SubEdits} 
SubEdits \citep{chollampatt-etal-2020-automatic} is collected from Rakuten Viki, a popular video streaming platform. 
Subtitle segments are clipped from videos, translated by a proprietary NMT system and post-edited by volunteers from the community when necessary. 
SubEdits is a large human-annotated APE dataset containing 161K subtitle domain triplets in total. 

\paragraph{MLQE-PE} 
Source texts in MLQE-PE \citep{2010.04480} are selected from Wikipedia and translated by an in-domain state-of-art NMT model. 
Human post-edits in MLQE-PE are generated by paid editors from the Unbabel community. 
MLQE-PE has been used in WMT 2021 APE shared task. 

\paragraph{eSCAPE} 
eSCAPE corpus \citep{1803.07274} is a large-scale artificial APE dataset constructed by direct translation. 
Parallel corpora from different domains are collected and merged to train a NMT model. 
Output of the NMT model is used as \textit{mt} and \textit{ref} in parallel corpora is used directly as \textit{pe} to compose a triplet. 

\paragraph{SubEscape} 
SubEscape \citep{chollampatt-etal-2020-automatic} is an artificial corpus that is constructed by the same method as eSCAPE but only contains subtitles domain triplets.

\subsection{Models}

Transformer-based models are proved to be advantageous in processing outputs from both SMT and NMT systems.
Most competitors in WMT 2020 APE task \citep{lee-2020-cross,lee-etal-2020-postech,lee-etal-2020-noising,wang-etal-2020-alibabas,yang-etal-2020-hw} and WMT 2021 APE task \citep{sharma-etal-2021-adapting, oh-etal-2021-netmarble} adopt models based on the Transformer architecture. 
Therefore, we choose two typical Transformer-based models, Transformer and \textit{BERT Enc.+BERT Dec.}, to conduct comparative experiments on.  
Both models are used in the comparison of construction methods, and we choose \textit{BERT-APE} for domain study.
\paragraph{Transformer} 
Transformer \citep{1706.03762} employs multi-head attention to extract features from the input sequence. 
It is widely used in natural language processing and has achieved relatively excellent performance on APE task.

\paragraph{\textit{BERT Enc.+BERT Dec. (BERT-APE)}} 
\textit{BERT Enc.+BERT Dec.} \citep{1906.06253}, (simplified as \textit{BERT-APE}), initializes decoder with parameters of BERT and shares the self-attention module in encoder and decoder. 
Since there is no context attention layer in BERT, \textit{BERT-APE} initializes context attention layers in decoder with the parameters of self-attention layer in BERT. \textit{BERT-APE} is now the state-of-art model on WMT'18 SMT and SubEdits.

We report our model configurations and their hyperparameters in Appendix~\ref{sec:configuration}. 
In all our experiments, APE models are firstly pre-trained with artificial triplets and then fine-tuned on genuine data.

\subsection{Evaluation}

Following previous research on APE, we evaluate the ouput of APE systems with three different automatic metrics, BLEU \citep{papineni-etal-2002-bleu}, ChrF  \citep{popovic-2015-chrf} and TER \citep{snover-etal-2006-study}. 
We compute BLEU and ChrF with SacreBLEU \citep{post-2018-call}, and TER with TERCOM. 

\section{Comparison of Construction Methods}

Considering the importance of data augmentation in APE, different construction methods of artificial triplets have been proposed. 
In this section, we mainly study three representative methods for creating artificial triplets to examine the importance of \textit{src, mt and pe} in synthetic data.

\subsection{Construction Methods}

\paragraph{Direct-Trans} 
\citet{1803.07274} created a widely-used artificial corpus, eSCAPE, by training a MT system to translate the source text (\textit{src}) of parallel corpora to obtain \textit{mt} and using \textit{ref} of parallel corpora as \textit{pe}. 
The strength of this method is that \textit{mt}, which is generated in the same way as genuine data, can reflect errors in real MT outputs. 
However, since the edit distance between \textit{mt} and the created \textit{pe} is larger than real human-edits, the correlation between \textit{mt} and \textit{pe} will be weaken. 
Overall, Direct-Trans constructs artificial triplets with high-quality \textit{src} and \textit{mt} but low-quality \textit{pe}.

\paragraph{Round-Trans} 
A competitor of the WMT 2016 APE task constructed a large artificial APE dataset with Round-Trans method \citep{1605.04800} . 
During the process, two MT models need to be trained, one from English to German and the other from German to English. 
Monolingual sentences are directly used as \textit{pe} and then translated by the two MT systems to obtain \textit{src} and \textit{mt}. 
Round-Trans ensures the quality of \textit{pe}, but the quality of \textit{src} and \textit{mt} generated by trained MT models are unstable.

\paragraph{Noising} 
Applying editing operations to smooth texts is a common data augmentation method in grammatical error correction \citep{awasthi-etal-2019-parallel}. \citet{lee-etal-2020-noising} creates artificial triplets by adding noise to \textit{ref} of parallel corpora to generate \textit{mt}. 
In our experiments, we randomly perform edit operations, including Random Insertion, Random Swap and Random Deletion \citep{wei-zou-2019-eda} to \textit{ref}. 
The probability of adding noise to each word is denoted by \textit{p}. 
We set \textit{p} to 0.05 and 0.1 respectively to study the influence of \textit{p} value.

\subsection{Experimental Results}

We use eSCAPE corpus as a representative of Direct-Trans. 
Edit operations are performed to \textit{pe} in eSCAPE to construct the Noising artificial data.
We take \textit{pe} in eSCAPE as monolingual corpus to implement Round-Trans method. 
Experimental results are shown in Table \ref{tab:method}.

\begin{table*}
\centering 
\begin{tabular}{|c|c|c|c|c|c|c|c|} 
\hline
 Model & Method & \multicolumn{3}{|c|}{SubEdits} &   \multicolumn{3}{|c|}{WMT'18 SMT}\\
 \hline
  \multicolumn{2}{|c|}{}  & BLEU↑ & ChrF↑ & TER↓ &  BLEU↑ & ChrF↑ & TER↓ \\
\hline
  \multicolumn{2}{|c|}{Without APE} & 61.9 & 71.3 & 27.3 & 63.4 & 82.5 & 23.6 \\
\hline
\multirowcell{5}{\textit{BERT-APE}} & Direct-Trans & \textbf{65.7}&\textbf{75.6} &\textbf{23.1} &\textbf{72.2} &\textbf{86.0} &\textbf{17.5} \\ 
\cline{2-8}
& Round-Trans &63.4 &73.4 &24.2 &70.7 &85.2 &18.6 \\
\cline{2-8}
& Noising(\textit{p}=0.05) &63.8 &73.8 &23.8 &69.7 &84.6 &19.4 \\
\cline{2-8}
& Noising(\textit{p}=0.1) &64.1 &74.2 &23.6 &70.4 &85.2 &18.6 \\
\hline
\multirowcell{5}{Transformer} &  Direct-Trans & {64.1}&{74.2} &{23.4} &{71.0} &{85.4} &{18.6} \\ 
\cline{2-8}
&Round-Trans &64.0 &74.2 &23.8 &70.5 &84.9 &18.6 \\
\cline{2-8}
&Noising(\textit{p}=0.05) & 63.1&73.1 &24.6 &68.5 &83.9 &19.9 \\
\cline{2-8}
&Noising(\textit{p}=0.1)&64.1& 74.1& 23.5& 69.3& 84.5&19.6  \\
\hline
\end{tabular}
 \\
\caption{\label{tab:method}
Performance of APE models pre-trained with artificial corpora constructed by Direct-Trans, Round-Trans and Noising.
}
\end{table*}

Seen from the table, artificial data constructed by Direct-Trans method help APE models achieve the best performance on both SubEdits and WMT'18 SMT. 
This result demonstrates that with \textit{src} and \textit{mt} closer to real outputs of MT systems, APE models are more likely to work better. 
Besides, when employing Noising method, changing the \textit{p} value from 0.05 to 0.1 actually improves the performance of both models. 
We suppose that the number of errors in \textit{mt} has nothing to do with the training of APE models. 
In this case, more noise in \textit{mt} does not result in a decrease in model performance, but actually benefits it.
One possible reason for the slightly poorer performance of Noising than Direct-Trans is that artificial triplets created by Noising can only cover limited error types.  
Besides, taking semantic features into account is difficult when corrupting \textit{ref}. 
Translation errors in relation to semantics are hard to be artificially constructed in this way.
In general, constructed artificial triplets are supposed to help APE models learn how to correct errors in \textit{mt}. 
Triplets constructed by Direct-Trans are able to cover more error types in \textit{mt}, thus closer to the distribution of genuine data.

\begin{table*}
\centering 
\begin{tabular}{c|ccc|ccc} 
\hline
 Domain & \multicolumn{3}{|c|}{SubEdits} &   \multicolumn{3}{|c}{WMT'18 SMT}\\
 \hline
    & BLEU↑ & ChrF↑ & TER↓ &  BLEU↑ & ChrF↑ & TER↓ \\
\hline
Without pre-train & 63.9 & 73.9 & 24.8 & 67.1 & 83.0 & 20.8 \\
 IT & 63.0& 73.1& 24.6& \textbf{68.9}${\dagger}$& \textbf{84.6}${\dagger}$ & \textbf{19.7}${\dagger}$ \\
 Legal & 63.5&\textbf{74.1} &24.6 &64.5 &81.9 &23.2 \\
 Medic & 62.9&72.7 &24.9 &65.3 &82.1 &22.6 \\
 News & \textbf{64.1} &74.0 &24.5 &55.1 &64.8 & 31.2\\
 Subtitles & \textbf{64.1}${\dagger}$ & \textbf{74.1}${\dagger}$ & \textbf{24.0}${\dagger}$ & 67.9&83.5 &20.5 \\
\hline
\end{tabular}
 \\
\caption{\label{tab:single_domain}
APE performance on SubEdits and WMT'18 SMT pre-trained with synthetic data from different domains. Values marked with ${\dagger}$ are results pre-trained with in-domain data. 
}
\end{table*}

\section{Comparison of Data Domain}

In view of the fact that NMT performance is particularly domain-dependant \citep{chu-wang-2018-survey}, \citet{chollampatt-etal-2020-automatic} empirically demonstrated that in-domain human post-edited data are more helpful to APE models than out-of-domain human post-edited triplets. 
However, no previous research has studied the influence of artificial data domain. 
In order to investigate it, we pre-train \textit{BERT-APE} with artificial data from different domains, and assess the model performance on WMT'18 SMT and SubEdits.

\begin{figure*} [h]
\centering
\begin{subfigure}[b]{0.45\textwidth}
    \centering
    \includegraphics[width=\textwidth]{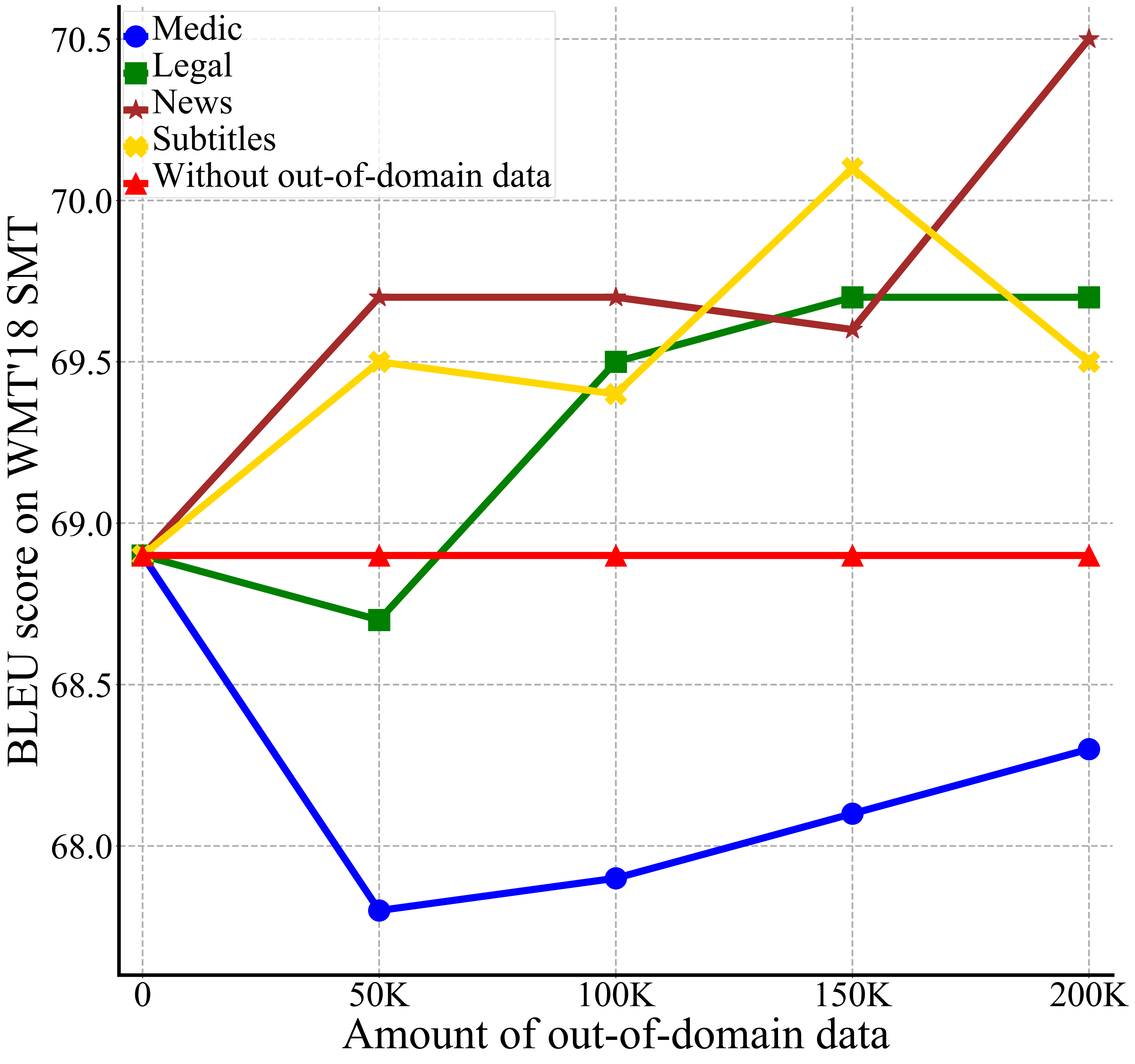}
    \caption{BLEU score of APE outputs on WMT'18 SMT}
    \label{sfig:domain_smt}
\end{subfigure}
\hfill
\begin{subfigure}[b]{0.45\textwidth}
    \centering
    \includegraphics[width=\textwidth]{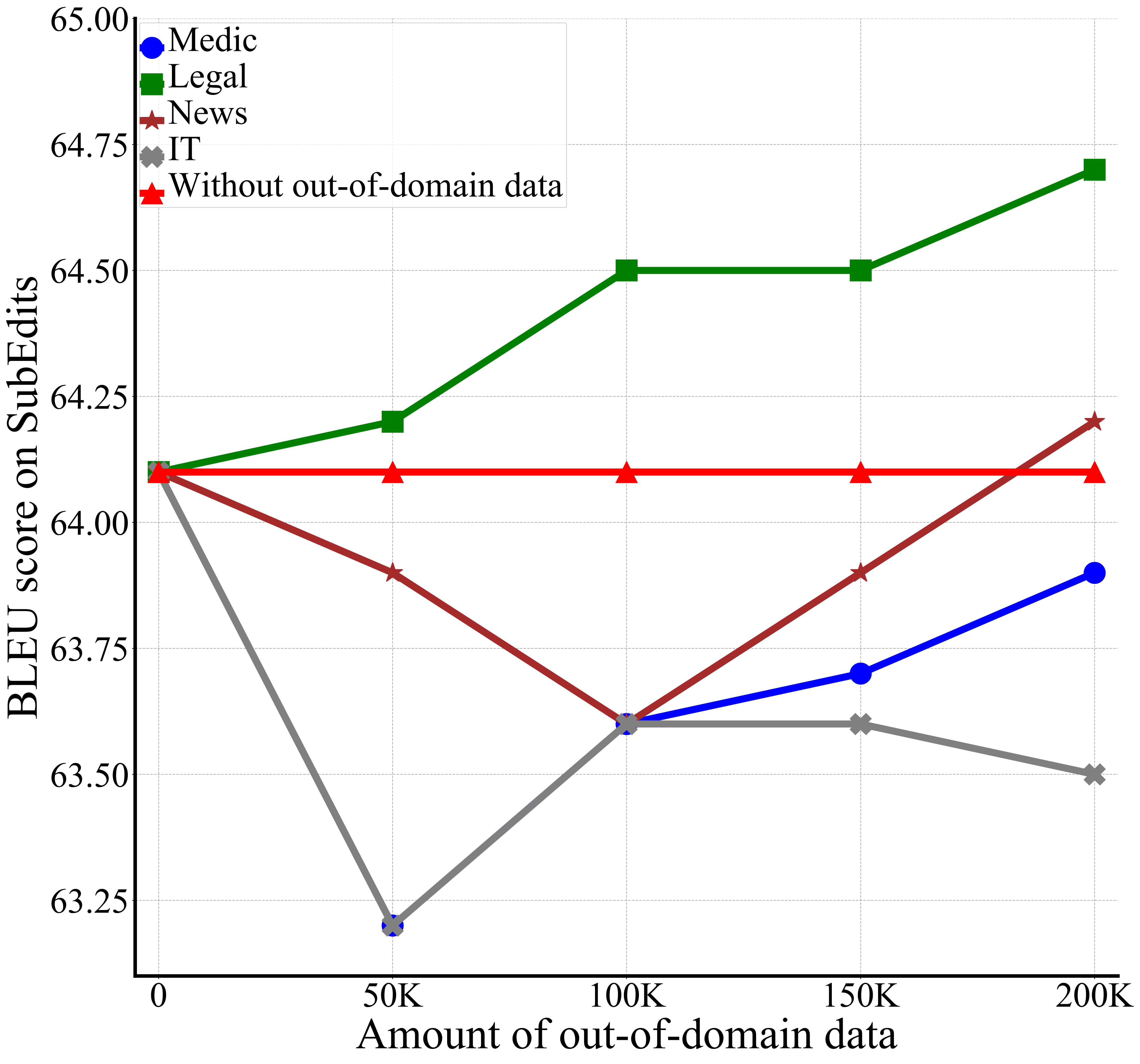}
    \caption{BLEU score of APE outputs on SubEdits}
    \label{sfig:domain_subedits}
\end{subfigure}
\caption{Trend of model performance when incorporating out-of-domain artificial data. Each curve in the figure illustrates model performance when incorporating data from a certain outside domain. The red line in the figure gives a baseline BLEU score obtained without incorporating any out-of-domain data.}
\label{fig:data_domain}
\end{figure*}

\subsection{In-domain v.s. Out-of-domain}

To compare in-domain and out-of-domain artificial triplets, we split the eSCAPE dataset into several small corpora that contain single-domain data from IT, Medic, Legal and News. 
To deal with the imbalance between domains, we control the amount of data from each domain equal, containing 200K training examples. 
The same amount of data are sampled from SubEscape to obtain single-domain data from subtitles. 
The results are reported in Table \ref{tab:single_domain}.

In-domain artificial data help the APE model achieve better results than out-domain synthetic data on both SubEdits and WMT'18 SMT. 
It can be found that performance of the APE model has declined when pre-trained with some single-domain data. 
For example, pre-training with IT and Medic data reduces the BLEU score on SubEdits. 
We speculate that out-of-domain artificial data from relevant domains can improve APE performance, while others may interfere with the training of the APE model. 

\subsection{What Can Out-of-domain Data Do?}

In practical applications, there might be a serious imbalance between artificial triplets from different domains. 
The amount of parallel datasets collected by eSCAPE corpus varies across domains. 
The amount of data from IT domain is only about one tenth of those from Medic domain. 
Therefore, considering the difficulty constructing large-scale artificial in-domain triplets, out-of-domain data is necessary when implementing APE on translation outputs. 
In addition, artificial data from similar domains can possibly help APE system learn more useful knowledge. 
We keep the in-domain subset unchanged with 200K training examples, and incorporate out-of-domain data into it to observe the effect of out-of-domain triplets. 
The trend is illustrated in Figure \ref{fig:data_domain} and complete experimental results are available in Appendix~\ref{sec:data_domain}.

As is shown in Figure \ref{sfig:domain_subedits}, contrary to expectations, incorporating data from Medic and IT domains does not improve the performance of the APE model, but results in a decrease in BLEU score. 
The experimental results indicate that synthetic triplets from irrelevant domains are not helpful to our APE model. 
A large-scale artificial APE datasets of mixed domains are likely to contain some triplets from irrelevant domains, which are harmful to the performance of our model.

\begin{table*}
\centering 
\begin{tabular}{c|ccc|ccc} 
\hline
 eSCAPE & \multicolumn{3}{|c|}{SubEdits} &   \multicolumn{3}{|c}{WMT'18 SMT}\\
 \hline
    & BLEU↑ & ChrF↑ & TER↓ &  BLEU↑ & ChrF↑ & TER↓ \\
\hline
Unfiltered & 65.7 & \textbf{75.6} & 23.1 & 72.2 & 86.0 & 17.5 \\
Filtered & \textbf{65.9} & 75.5 & \textbf{22.8} & \textbf{72.8} & \textbf{86.4} & \textbf{17.1} \\
\hline
\end{tabular}
 \\
\caption{\label{tab:filtered}
The performance of \textit{BERT-APE} model pre-trained with eSCAPE corpus unfiltered and filtered out Medic data.
}
\end{table*}

\subsection{Rethinking of Synthetic Data Domain}

Our experiments illustrate that artificial triplets from inappropriate domains are harmful to the performance of the APE model. 
Some researches \citep{lee-etal-2020-noising,lee-etal-2021-adaptation} that work from the data perspective, have proposed that low-quality training data can interfere on APE models. 
We have found that in addition to low-quality artificial triplets, data from unrelated domains will also do harm to the APE model. 
As shown in Figure \ref{fig:data_domain}, Medic synthetic triplets are not helpful both on WMT'18 and SubEdits.
Therefore, we hypothesize when performing APE on the translation results of a specific domain, filtering out artificial data from irrelevant domains can not only speed up the convergence of the model, but also improve the final performance of the system. 

Motivated by this idea, we filter out Medic data from eSCAPE and pre-train the \textit{BERT-APE} model with the rest artificial triplets. 
The experimental results are shown in Table \ref{tab:filtered}. 
The APE model has done a better job on both SubEdits and WMT'18 SMT. 
From this point of view, when applying large-scale artificial datasets like eSCAPE, domain filtering might be a common way to help improve model outputs.

\section{Problem Analysis}

Although existing APE models work well on some APE datasets, they fail to achieve the same improvement on outputs of strong NMT systems. 
Our method of pre-training \textit{BERT-APE} model on filtered eSCAPE corpus and then fine-tuning on genuine data has achieved an improvement of +4.0 BLEU and -4.5 TER on SubEdits, and +9.4 BLEU and -5.5 TER on WMT'18 SMT, but results in a decrease of -0.9 BLEU and +0.8 TER on MLQE-PE.
In WMT 2021 APE shared task, the best result of the contestants has achieved an improvement of +0.46 BLEU and -0.77 TER, which is much worse than the performance on SubEdits and WMT'18 SMT.  
Therefore, in this section, we aim to analyze the reasons for the poor performance on MLQE-PE and explore the problems of current APE systems.
Although MLQE-PE is a mixed-domain dataset, examples from Medic domain are only a very small part of it. 
Consequently, we filter out the Medic data from eSCAPE as well.
We conduct research on the output of \textit{BERT-APE} pre-trained on filtered eSCAPE, which achieves the best results in our experiments. 

\subsection{Impact of Text Length}

It is known that large language models tend to generate incoherent long texts \citep{2203.11370}. 
In order to study the impact of this problem on the performance of APE systems, we split the test set of WMT'18 SMT, SubEdits and MLQE-PE into 4 subsets according to the length of \textit{src}. 
We use $\Delta$BLEU, the improvement of BLEU score from \textit{mt} to APE output, to measure the effect of APE.   

\begin{figure}[t]
\centering
\includegraphics[width=0.45\textwidth]{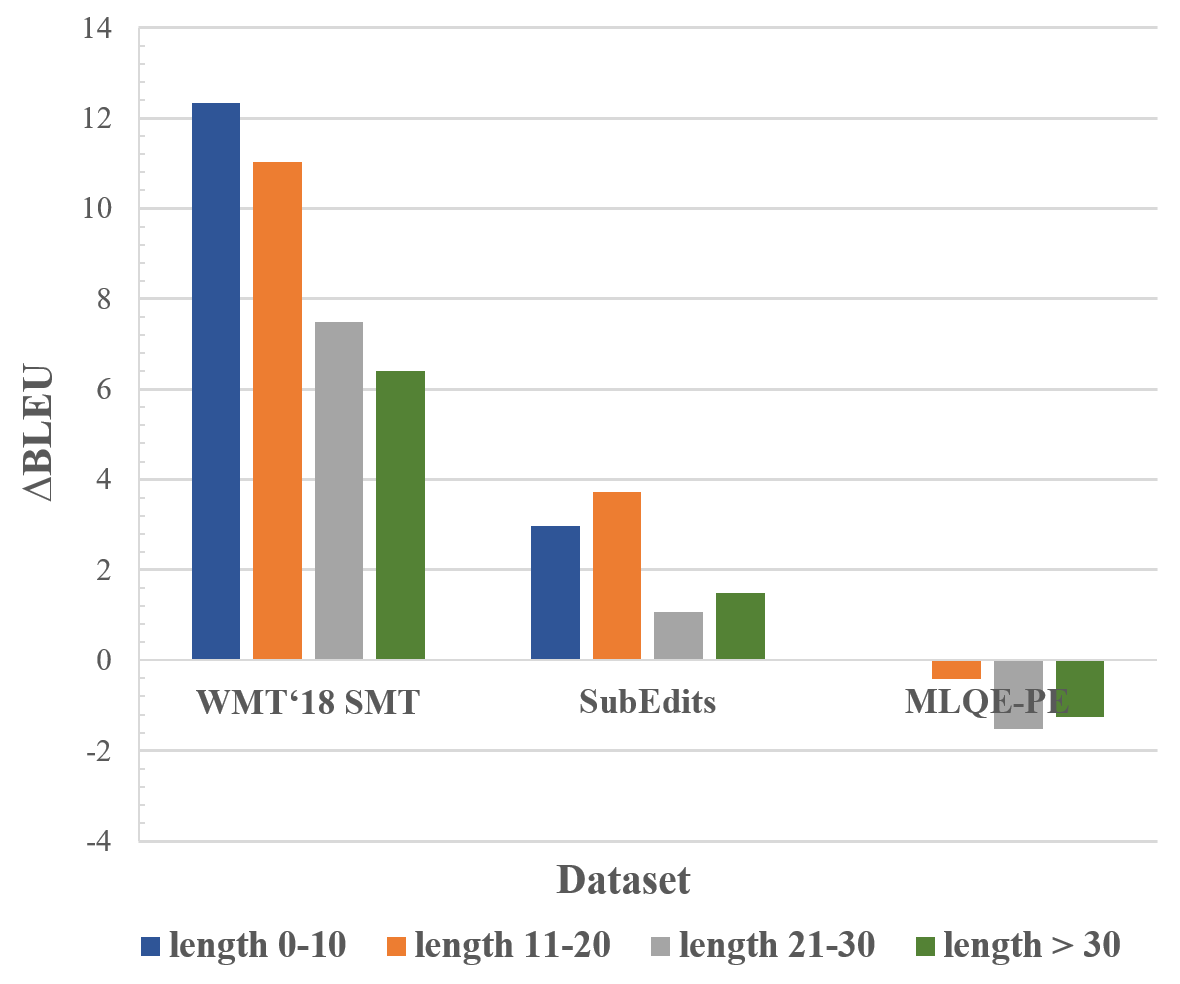}
\caption{APE improvement w.r.t. different lengths of \textit{src}. Negative $\Delta$BLEU indicates APE output is even worse than original \textit{mt}.}
\label{fig:APE_len}
\end{figure}

As is shown in Figure \ref{fig:APE_len}, as the text length increases, the performance of the APE model declines in WMT'18 SMT, and has a declining trend in the other datasets. 
Considering BERT is applied in the APE model, we speculate that the problem of long text generation might also exist in this APE system.
Detailed statistics can be found in Appendix~\ref{sec:statistics}.

\subsection{Impact of Translation Quality}

\begin{figure}[t]
\centering
\includegraphics[width=0.45\textwidth]{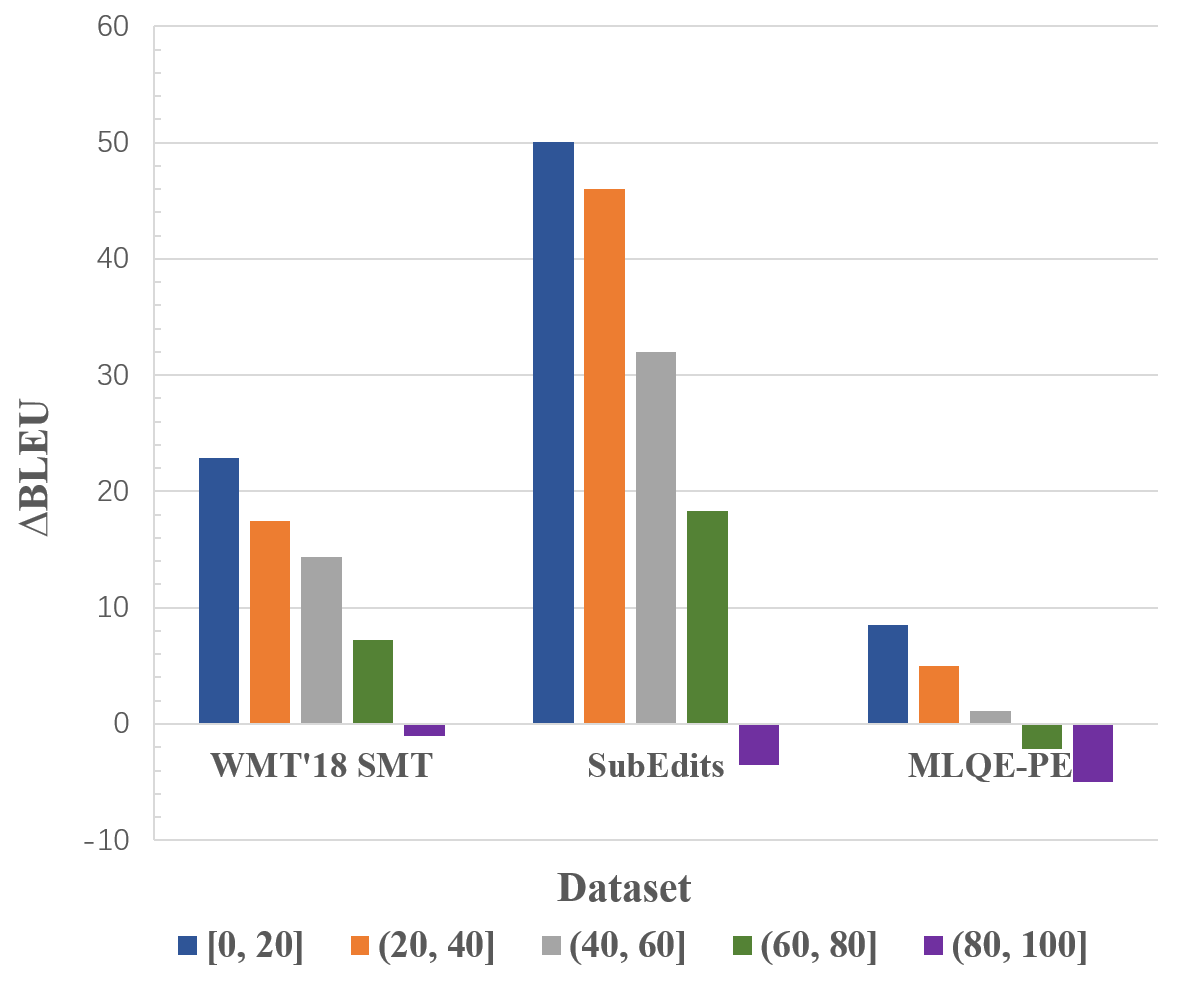}
\caption{APE improvement w.r.t. varying quality of \textit{mt}.}
\label{fig:APE_quality}
\end{figure}

To study the impact of APE with varying quality of MT output, we conduct analysis on test sets of the three APE datasets (Figure \ref{fig:APE_quality}). 
We split the dataset into 5 subsets by aggregating triplets with \textit{mt-pe} sentence BLEU in [0,20], (20,40], (40,60], (60,80] and (80,100], respectively. 
Whether for SMT or NMT output, the APE system performs poorly on high-quality \textit{mt}, while it does a good job on low-quality \textit{mt}. 
For translation results with sentence BLEU exceeding 80, the APE model will reduce the translation quality on all three datasets. 
We suppose there are only some subtle errors in these high-quality MT outputs, but the APE system modifies some correctly translated-texts, resulting in poor performance on these data. 
Nearly half of \textit{mt} in MLQE-PE have sentence BLEU score over 80, which can be one of the reasons why the APE model performs the worst on this dataset. 
Therefore, we believe that APE models need to learn more about which translation outputs need modifying and which ones are already correct. 
When employing post-editing on \textit{mt}, the APE model is supposed to estimate the quality of \textit{mt} and vary its cautiousness.

\subsection{Human Annotation}

\begin{table*}
\centering 
\begin{tabular}{c|c|c|c|c|c} 
\hline
Error type & in \textit{mt} & corrected in APE & new in APE & corrected in \textit{pe} &  new in \textit{pe} \\
\hline
Omission & 3 & 0 & 10 & 3 & 0 \\
Addition & 7 & 6 & 0 & 4 &3 \\
Entity error & 12 & 4 & 18 & 7 & 0 \\
Polysemy error & 8 & 2 & 1 & 2 & 2 \\
Word Order & 6 & 2 & 0 & 2 & 4 \\
Grammatical error & 25 & 12 & 0 & 17 & 4 \\
Translation word error & 47 & 4 & 21 & 12 & 13 \\ 
\hline
\end{tabular}
\caption{\label{tab:human_annotation}
Number of various error types in the 100 triplets used for human annotation. In addition to counting the number of \textit{mt} containing each type of error, we further investigate the number of examples APE and \textit{pe} correct the error in \textit{mt} and the number of examples APE and \textit{pe} produce new errors.
}
\end{table*}

There are different translation problems in MT outputs. 
We believe that it is necessary to analyze what problems Transformer-based Seq2Seq models can solve and what they cannot. 
Most of the existing APE models adopt the traditional Seq2Seq paradigm without dealing with specific types of problems. 
Understanding the problems of the APE models not only allows for better tuning of the model structure, but also helps the APE system assist humans in post-editing MT outputs. 
We classify errors in translation results into 7 categories and conduct human annotation on error types of \textit{mt}, \textit{pe} and APE output. 
Our human annotators are all college students majoring in German and they have passed English proficiency examinations.
They know common English and German words well and are familiar with the grammatical requirements of English and German.
Before formal annotation, an instruction document with examples of different error types is provided to help them understand various error types clearly.
We select 100 triplets from MLQE-PE test set, and ask annotators to respectively evaluate \textit{mt}, APE output, and \textit{pe} as a German translation of \textit{src} (Table \ref{tab:human_annotation}). 
They need to point out types of errors in the translation according to \textit{src}. 
Since one translated-text may contain various kinds of errors, annotators can note multiple error types for one translation. 
Each text is annotated by two annotators independently.
If their results differ, a third annotator is responsible for determining the final result.
The errors we define include the following types:

\paragraph{Omission} The translated-text misses some semantic information in the source text.

\paragraph{Addition} The translated-text contains semantic information not found in the source text.

\paragraph{Entity error} Incorrect translation of name entities in the source text.

\paragraph{Polysemy error} Some word in the source text has multiple meanings and a wrong meaning is chosen in the translated text.

\paragraph{Word Order} Improper word order in translated-text.

\paragraph{Grammatical error} Problems with tense, word form and some other grammatical errors in the translated-text.

\paragraph{Translation word error} Although the meaning of the source text is translated correctly, the translated-text is not fluent due to the problem of word selection in translation. 

Through comparing the results of APE output and \textit{pe} on Omission and Addition, we can see that as opposed to human post-editing, the APE system has an noticeable effect on solving Addition problems, but fails on Omission. 
APE output is not likely to contain semantic information not appearing in \textit{src}, but some important information that have been translated in \textit{mt} might get lost. Moreover, the APE model has a relatively good effect on solving grammatical errors in \textit{mt}. 
The output of the model correctly revises grammatical problems in half of the examples without generating new grammatical errors. 
However, the APE system works extremely terrible at handling entity errors. 
Not only does the APE model fail to correct the translation errors of name entities in most \textit{mt}, it rather generates more mistranslations of name entities. 
This may be related to the lack of corresponding information between name entities in \textit{src} and \textit{mt} during the training process. 
The model does not learn how to recognize name entities in \textit{src} and \textit{mt}, or how to translate entities into target language. 
Therefore, we suppose that adding entity information in the training process can help the model perform better.

It is not easy to obtain a fluent translation by correcting errors in the output of MT system, as there are some problems in the text structure and word selection of \textit{mt} that are hard to address by text-editing. 
Even human post-editing cannot resolve translation word error well. 
Nevertheless, entity error and omission that will affect the semantics of the APE output, are more worthy of attention. 

\section{Conclusion}

We empirically compare construction methods of artificial training data and data domain of synthetic data on Transformer-based APE models. 
We draw several conclusions to help construct better artificial training triplets in the future, for example, domain filtering can be helpful. 
We also analyze the problems of the state-of-art APE model on MLQE-PE dataset. 
Our work shows that although the APE model has achieved good results on some MT outputs, it fails on addition and entity error, which is worth exploring in the future.

\section{Limitations}

We find that some out-of-domain artificial data will negatively affect the automatic post-editing systems. 
However, we fail to give a strict answer to the question that what kinds of domains can help improve the performance.
Further research and experiments are needed to find a more accurate and effective data filtering method to exclude artificial triplets that are making interference.
Besides, our experiments only focus on APE in en-de. 
Multilingual APE task is also worthy of future research.

\bibliography{anthology,custom}

\begin{thebibliography}{38}
\expandafter\ifx\csname natexlab\endcsname\relax\def\natexlab#1{#1}\fi

\bibitem[{Awasthi et~al.(2019)Awasthi, Sarawagi, Goyal, Ghosh, and
  Piratla}]{awasthi-etal-2019-parallel}
Abhijeet Awasthi, Sunita Sarawagi, Rasna Goyal, Sabyasachi Ghosh, and Vihari
  Piratla. 2019.
\newblock \href {https://doi.org/10.18653/v1/D19-1435} {Parallel iterative edit
  models for local sequence transduction}.
\newblock In \emph{Proceedings of the 2019 Conference on Empirical Methods in
  Natural Language Processing and the 9th International Joint Conference on
  Natural Language Processing (EMNLP-IJCNLP)}, pages 4260--4270, Hong Kong,
  China. Association for Computational Linguistics.

\bibitem[{Bojar et~al.(2017)Bojar, Chatterjee, Federmann, Graham, Haddow,
  Huang, Huck, Koehn, Liu, Logacheva, Monz, Negri, Post, Rubino, Specia, and
  Turchi}]{bojar-etal-2017-findings}
Ond{\v{r}}ej Bojar, Rajen Chatterjee, Christian Federmann, Yvette Graham, Barry
  Haddow, Shujian Huang, Matthias Huck, Philipp Koehn, Qun Liu, Varvara
  Logacheva, Christof Monz, Matteo Negri, Matt Post, Raphael Rubino, Lucia
  Specia, and Marco Turchi. 2017.
\newblock \href {https://doi.org/10.18653/v1/W17-4717} {Findings of the 2017
  conference on machine translation ({WMT}17)}.
\newblock In \emph{Proceedings of the Second Conference on Machine
  Translation}, pages 169--214, Copenhagen, Denmark. Association for
  Computational Linguistics.

\bibitem[{Chatterjee et~al.(2019)Chatterjee, Federmann, Negri, and
  Turchi}]{chatterjee-etal-2019-findings}
Rajen Chatterjee, Christian Federmann, Matteo Negri, and Marco Turchi. 2019.
\newblock \href {https://doi.org/10.18653/v1/W19-5402} {Findings of the {WMT}
  2019 shared task on automatic post-editing}.
\newblock In \emph{Proceedings of the Fourth Conference on Machine Translation
  (Volume 3: Shared Task Papers, Day 2)}, pages 11--28, Florence, Italy.
  Association for Computational Linguistics.

\bibitem[{Chatterjee et~al.(2018)Chatterjee, Negri, Rubino, and
  Turchi}]{chatterjee-etal-2018-findings}
Rajen Chatterjee, Matteo Negri, Raphael Rubino, and Marco Turchi. 2018.
\newblock \href {https://doi.org/10.18653/v1/W18-6452} {Findings of the {WMT}
  2018 shared task on automatic post-editing}.
\newblock In \emph{Proceedings of the Third Conference on Machine Translation:
  Shared Task Papers}, pages 710--725, Belgium, Brussels. Association for
  Computational Linguistics.

\bibitem[{Chollampatt et~al.(2020)Chollampatt, Susanto, Tan, and
  Szymanska}]{chollampatt-etal-2020-automatic}
Shamil Chollampatt, Raymond~Hendy Susanto, Liling Tan, and Ewa Szymanska. 2020.
\newblock \href {https://doi.org/10.18653/v1/2020.emnlp-main.217} {Can
  automatic post-editing improve {NMT}?}
\newblock In \emph{Proceedings of the 2020 Conference on Empirical Methods in
  Natural Language Processing (EMNLP)}, pages 2736--2746, Online. Association
  for Computational Linguistics.

\bibitem[{Chu and Wang(2018)}]{chu-wang-2018-survey}
Chenhui Chu and Rui Wang. 2018.
\newblock \href {https://aclanthology.org/C18-1111} {A survey of domain
  adaptation for neural machine translation}.
\newblock In \emph{Proceedings of the 27th International Conference on
  Computational Linguistics}, pages 1304--1319, Santa Fe, New Mexico, USA.
  Association for Computational Linguistics.

\bibitem[{Correia and Martins(2019)}]{1906.06253}
Gonçalo~M. Correia and André F.~T. Martins. 2019.
\newblock \href {http://arxiv.org/abs/arXiv:1906.06253} {A simple and effective
  approach to automatic post-editing with transfer learning}.

\bibitem[{Devlin et~al.(2018)Devlin, Chang, Lee, and Toutanova}]{1810.04805}
Jacob Devlin, Ming-Wei Chang, Kenton Lee, and Kristina Toutanova. 2018.
\newblock \href {http://arxiv.org/abs/arXiv:1810.04805} {Bert: Pre-training of
  deep bidirectional transformers for language understanding}.

\bibitem[{Fomicheva et~al.(2020)Fomicheva, Sun, Fonseca, Zerva, Blain,
  Chaudhary, Guzmán, Lopatina, Specia, and Martins}]{2010.04480}
Marina Fomicheva, Shuo Sun, Erick Fonseca, Chrysoula Zerva, Frédéric Blain,
  Vishrav Chaudhary, Francisco Guzmán, Nina Lopatina, Lucia Specia, and André
  F.~T. Martins. 2020.
\newblock \href {http://arxiv.org/abs/arXiv:2010.04480} {Mlqe-pe: A
  multilingual quality estimation and post-editing dataset}.

\bibitem[{Grangier and Auli(2018)}]{grangier-auli-2018-quickedit}
David Grangier and Michael Auli. 2018.
\newblock \href {https://doi.org/10.18653/v1/N18-1025} {{Q}uick{E}dit: Editing
  text {\&} translations by crossing words out}.
\newblock In \emph{Proceedings of the 2018 Conference of the North {A}merican
  Chapter of the Association for Computational Linguistics: Human Language
  Technologies, Volume 1 (Long Papers)}, pages 272--282, New Orleans,
  Louisiana. Association for Computational Linguistics.

\bibitem[{Huang et~al.(2019)Huang, Liu, Luan, Xu, and
  Sun}]{huang-etal-2019-learning-copy}
Xuancheng Huang, Yang Liu, Huanbo Luan, Jingfang Xu, and Maosong Sun. 2019.
\newblock \href {https://doi.org/10.18653/v1/D19-1634} {Learning to copy for
  automatic post-editing}.
\newblock In \emph{Proceedings of the 2019 Conference on Empirical Methods in
  Natural Language Processing and the 9th International Joint Conference on
  Natural Language Processing (EMNLP-IJCNLP)}, pages 6122--6132, Hong Kong,
  China. Association for Computational Linguistics.

\bibitem[{Ive et~al.(2020)Ive, Specia, Szoc, Vanallemeersch, Van~den Bogaert,
  Farah, Maroti, Ventura, and Khalilov}]{ive-etal-2020-post}
Julia Ive, Lucia Specia, Sara Szoc, Tom Vanallemeersch, Joachim Van~den
  Bogaert, Eduardo Farah, Christine Maroti, Artur Ventura, and Maxim Khalilov.
  2020.
\newblock \href {https://aclanthology.org/2020.lrec-1.455} {A post-editing
  dataset in the legal domain: Do we underestimate neural machine translation
  quality?}
\newblock In \emph{Proceedings of the 12th Language Resources and Evaluation
  Conference}, pages 3692--3697, Marseille, France. European Language Resources
  Association.

\bibitem[{Junczys-Dowmunt and Grundkiewicz(2016)}]{1605.04800}
Marcin Junczys-Dowmunt and Roman Grundkiewicz. 2016.
\newblock \href {http://arxiv.org/abs/arXiv:1605.04800} {Log-linear
  combinations of monolingual and bilingual neural machine translation models
  for automatic post-editing}.

\bibitem[{Junczys-Dowmunt and
  Grundkiewicz(2017)}]{junczys-dowmunt-grundkiewicz-2017-exploration}
Marcin Junczys-Dowmunt and Roman Grundkiewicz. 2017.
\newblock \href {https://aclanthology.org/I17-1013} {An exploration of neural
  sequence-to-sequence architectures for automatic post-editing}.
\newblock In \emph{Proceedings of the Eighth International Joint Conference on
  Natural Language Processing (Volume 1: Long Papers)}, pages 120--129, Taipei,
  Taiwan. Asian Federation of Natural Language Processing.

\bibitem[{Junczys-Dowmunt and Grundkiewicz(2018)}]{1809.00188}
Marcin Junczys-Dowmunt and Roman Grundkiewicz. 2018.
\newblock \href {http://arxiv.org/abs/arXiv:1809.00188} {Ms-uedin submission to
  the wmt2018 ape shared task: Dual-source transformer for automatic
  post-editing}.

\bibitem[{Lee(2020)}]{lee-2020-cross}
Dongjun Lee. 2020.
\newblock \href {https://aclanthology.org/2020.wmt-1.81} {Cross-lingual
  transformers for neural automatic post-editing}.
\newblock In \emph{Proceedings of the Fifth Conference on Machine Translation},
  pages 772--776, Online. Association for Computational Linguistics.

\bibitem[{Lee et~al.(2020{\natexlab{a}})Lee, Lee, Shin, Jung, Kim, and
  Lee}]{lee-etal-2020-postech}
Jihyung Lee, WonKee Lee, Jaehun Shin, Baikjin Jung, Young-Kil Kim, and
  Jong-Hyeok Lee. 2020{\natexlab{a}}.
\newblock \href {https://aclanthology.org/2020.wmt-1.82} {{POSTECH}-{ETRI}{'}s
  submission to the {WMT}2020 {APE} shared task: Automatic post-editing with
  cross-lingual language model}.
\newblock In \emph{Proceedings of the Fifth Conference on Machine Translation},
  pages 777--782, Online. Association for Computational Linguistics.

\bibitem[{Lee et~al.(2022)Lee, Heo, Jung, and Lee}]{2204.03896}
WonKee Lee, Seong-Hwan Heo, Baikjin Jung, and Jong-Hyeok Lee. 2022.
\newblock \href {http://arxiv.org/abs/arXiv:2204.03896} {Towards
  semi-supervised learning of automatic post-editing: Data-synthesis by
  infilling mask with erroneous tokens}.

\bibitem[{Lee et~al.(2021)Lee, Jung, Shin, and Lee}]{lee-etal-2021-adaptation}
WonKee Lee, Baikjin Jung, Jaehun Shin, and Jong-Hyeok Lee. 2021.
\newblock \href {https://doi.org/10.18653/v1/2021.eacl-main.322} {Adaptation of
  back-translation to automatic post-editing for synthetic data generation}.
\newblock In \emph{Proceedings of the 16th Conference of the European Chapter
  of the Association for Computational Linguistics: Main Volume}, pages
  3685--3691, Online. Association for Computational Linguistics.

\bibitem[{Lee et~al.(2020{\natexlab{b}})Lee, Shin, Jung, Lee, and
  Lee}]{lee-etal-2020-noising}
WonKee Lee, Jaehun Shin, Baikjin Jung, Jihyung Lee, and Jong-Hyeok Lee.
  2020{\natexlab{b}}.
\newblock \href {https://aclanthology.org/2020.wmt-1.83} {Noising scheme for
  data augmentation in automatic post-editing}.
\newblock In \emph{Proceedings of the Fifth Conference on Machine Translation},
  pages 783--788, Online. Association for Computational Linguistics.

\bibitem[{Negri et~al.(2018)Negri, Turchi, Chatterjee, and
  Bertoldi}]{1803.07274}
Matteo Negri, Marco Turchi, Rajen Chatterjee, and Nicola Bertoldi. 2018.
\newblock \href {http://arxiv.org/abs/arXiv:1803.07274} {escape: a large-scale
  synthetic corpus for automatic post-editing}.

\bibitem[{Oh et~al.(2021)Oh, Jang, Xu, An, and Oh}]{oh-etal-2021-netmarble}
Shinhyeok Oh, Sion Jang, Hu~Xu, Shounan An, and Insoo Oh. 2021.
\newblock \href {https://aclanthology.org/2021.wmt-1.34} {Netmarble {AI}
  center{'}s {WMT}21 automatic post-editing shared task submission}.
\newblock In \emph{Proceedings of the Sixth Conference on Machine Translation},
  pages 307--314, Online. Association for Computational Linguistics.

\bibitem[{Pal et~al.(2017)Pal, Naskar, Vela, Liu, and van
  Genabith}]{pal-etal-2017-neural}
Santanu Pal, Sudip~Kumar Naskar, Mihaela Vela, Qun Liu, and Josef van Genabith.
  2017.
\newblock \href {https://aclanthology.org/E17-2056} {Neural automatic
  post-editing using prior alignment and reranking}.
\newblock In \emph{Proceedings of the 15th Conference of the {E}uropean Chapter
  of the Association for Computational Linguistics: Volume 2, Short Papers},
  pages 349--355, Valencia, Spain. Association for Computational Linguistics.

\bibitem[{Pal et~al.(2016)Pal, Naskar, Vela, and van
  Genabith}]{pal-etal-2016-neural}
Santanu Pal, Sudip~Kumar Naskar, Mihaela Vela, and Josef van Genabith. 2016.
\newblock \href {https://doi.org/10.18653/v1/P16-2046} {A neural network based
  approach to automatic post-editing}.
\newblock In \emph{Proceedings of the 54th Annual Meeting of the Association
  for Computational Linguistics (Volume 2: Short Papers)}, pages 281--286,
  Berlin, Germany. Association for Computational Linguistics.

\bibitem[{Papineni et~al.(2002)Papineni, Roukos, Ward, and
  Zhu}]{papineni-etal-2002-bleu}
Kishore Papineni, Salim Roukos, Todd Ward, and Wei-Jing Zhu. 2002.
\newblock \href {https://doi.org/10.3115/1073083.1073135} {{B}leu: a method for
  automatic evaluation of machine translation}.
\newblock In \emph{Proceedings of the 40th Annual Meeting of the Association
  for Computational Linguistics}, pages 311--318, Philadelphia, Pennsylvania,
  USA. Association for Computational Linguistics.

\bibitem[{Popovi{\'c}(2015)}]{popovic-2015-chrf}
Maja Popovi{\'c}. 2015.
\newblock \href {https://doi.org/10.18653/v1/W15-3049} {chr{F}: character
  n-gram {F}-score for automatic {MT} evaluation}.
\newblock In \emph{Proceedings of the Tenth Workshop on Statistical Machine
  Translation}, pages 392--395, Lisbon, Portugal. Association for Computational
  Linguistics.

\bibitem[{Post(2018)}]{post-2018-call}
Matt Post. 2018.
\newblock \href {https://doi.org/10.18653/v1/W18-6319} {A call for clarity in
  reporting {BLEU} scores}.
\newblock In \emph{Proceedings of the Third Conference on Machine Translation:
  Research Papers}, pages 186--191, Brussels, Belgium. Association for
  Computational Linguistics.

\bibitem[{Sachan and Neubig(2018)}]{sachan-neubig-2018-parameter}
Devendra Sachan and Graham Neubig. 2018.
\newblock \href {https://doi.org/10.18653/v1/W18-6327} {Parameter sharing
  methods for multilingual self-attentional translation models}.
\newblock In \emph{Proceedings of the Third Conference on Machine Translation:
  Research Papers}, pages 261--271, Brussels, Belgium. Association for
  Computational Linguistics.

\bibitem[{Sennrich et~al.(2016)Sennrich, Haddow, and
  Birch}]{sennrich-etal-2016-improving}
Rico Sennrich, Barry Haddow, and Alexandra Birch. 2016.
\newblock \href {https://doi.org/10.18653/v1/P16-1009} {Improving neural
  machine translation models with monolingual data}.
\newblock In \emph{Proceedings of the 54th Annual Meeting of the Association
  for Computational Linguistics (Volume 1: Long Papers)}, pages 86--96, Berlin,
  Germany. Association for Computational Linguistics.

\bibitem[{Sharma et~al.(2021)Sharma, Gupta, and
  Nelakanti}]{sharma-etal-2021-adapting}
Abhishek Sharma, Prabhakar Gupta, and Anil Nelakanti. 2021.
\newblock \href {https://aclanthology.org/2021.wmt-1.35} {Adapting neural
  machine translation for automatic post-editing}.
\newblock In \emph{Proceedings of the Sixth Conference on Machine Translation},
  pages 315--319, Online. Association for Computational Linguistics.

\bibitem[{Simard et~al.(2007)Simard, Ueffing, Isabelle, and
  Kuhn}]{simard-etal-2007-rule}
Michel Simard, Nicola Ueffing, Pierre Isabelle, and Roland Kuhn. 2007.
\newblock \href {https://aclanthology.org/W07-0728} {Rule-based translation
  with statistical phrase-based post-editing}.
\newblock In \emph{Proceedings of the Second Workshop on Statistical Machine
  Translation}, pages 203--206, Prague, Czech Republic. Association for
  Computational Linguistics.

\bibitem[{Snover et~al.(2006)Snover, Dorr, Schwartz, Micciulla, and
  Makhoul}]{snover-etal-2006-study}
Matthew Snover, Bonnie Dorr, Rich Schwartz, Linnea Micciulla, and John Makhoul.
  2006.
\newblock \href {https://aclanthology.org/2006.amta-papers.25} {A study of
  translation edit rate with targeted human annotation}.
\newblock In \emph{Proceedings of the 7th Conference of the Association for
  Machine Translation in the Americas: Technical Papers}, pages 223--231,
  Cambridge, Massachusetts, USA. Association for Machine Translation in the
  Americas.

\bibitem[{Vaswani et~al.(2017)Vaswani, Shazeer, Parmar, Uszkoreit, Jones,
  Gomez, Kaiser, and Polosukhin}]{1706.03762}
Ashish Vaswani, Noam Shazeer, Niki Parmar, Jakob Uszkoreit, Llion Jones,
  Aidan~N. Gomez, Lukasz Kaiser, and Illia Polosukhin. 2017.
\newblock \href {http://arxiv.org/abs/arXiv:1706.03762} {Attention is all you
  need}.

\bibitem[{Vu and Haffari(2018)}]{vu-haffari-2018-automatic}
Thuy-Trang Vu and Gholamreza Haffari. 2018.
\newblock \href {https://doi.org/10.18653/v1/D18-1341} {Automatic post-editing
  of machine translation: A neural programmer-interpreter approach}.
\newblock In \emph{Proceedings of the 2018 Conference on Empirical Methods in
  Natural Language Processing}, pages 3048--3053, Brussels, Belgium.
  Association for Computational Linguistics.

\bibitem[{Wang et~al.(2020)Wang, Wang, Fan, Zhang, Lu, Ge, Shi, and
  Zhao}]{wang-etal-2020-alibabas}
Jiayi Wang, Ke~Wang, Kai Fan, Yuqi Zhang, Jun Lu, Xin Ge, Yangbin Shi, and
  Yu~Zhao. 2020.
\newblock \href {https://aclanthology.org/2020.wmt-1.84} {{A}libaba{'}s
  submission for the {WMT} 2020 {APE} shared task: Improving automatic
  post-editing with pre-trained conditional cross-lingual {BERT}}.
\newblock In \emph{Proceedings of the Fifth Conference on Machine Translation},
  pages 789--796, Online. Association for Computational Linguistics.

\bibitem[{Wang et~al.(2022)Wang, Durmus, Goodman, and Hashimoto}]{2203.11370}
Rose~E Wang, Esin Durmus, Noah Goodman, and Tatsunori Hashimoto. 2022.
\newblock \href {http://arxiv.org/abs/arXiv:2203.11370} {Language modeling via
  stochastic processes}.

\bibitem[{Wei and Zou(2019)}]{wei-zou-2019-eda}
Jason Wei and Kai Zou. 2019.
\newblock \href {https://doi.org/10.18653/v1/D19-1670} {{EDA}: Easy data
  augmentation techniques for boosting performance on text classification
  tasks}.
\newblock In \emph{Proceedings of the 2019 Conference on Empirical Methods in
  Natural Language Processing and the 9th International Joint Conference on
  Natural Language Processing (EMNLP-IJCNLP)}, pages 6382--6388, Hong Kong,
  China. Association for Computational Linguistics.

\bibitem[{Yang et~al.(2020)Yang, Wang, Wei, Shang, Guo, Li, Lei, Qin, Tao, Sun,
  and Chen}]{yang-etal-2020-hw}
Hao Yang, Minghan Wang, Daimeng Wei, Hengchao Shang, Jiaxin Guo, Zongyao Li,
  Lizhi Lei, Ying Qin, Shimin Tao, Shiliang Sun, and Yimeng Chen. 2020.
\newblock \href {https://aclanthology.org/2020.wmt-1.85} {{HW}-{TSC}{'}s
  participation at {WMT} 2020 automatic post editing shared task}.
\newblock In \emph{Proceedings of the Fifth Conference on Machine Translation},
  pages 797--802, Online. Association for Computational Linguistics.

\end{thebibliography}
\bibliographystyle{acl_natbib}

\clearpage
\appendix

\section{Model Configuration}
\label{sec:configuration}

Our model configurations are shown in Table \ref{tab:configuration}.

\begin{table} [htbp] 
\centering
\begin{tabular}{c|c|c} 
\hline
Settings & Transformer & \textit{BERT-APE}\\
\hline
Optimizer & Adam & AdamW  \\
Layers & 6 & 12 \\
Heads & 8 & 12\\
Hidden-size & 512 & 768 \\
Feed-forward & 2048 & 3072 \\
Batch-size & 4096 & 512 \\
Activation & RELU & GELU \\
Warmup steps & 5000 & 5000\\
Decay function & Noam & Noam \\
Training steps & 200K & 200K \\
Training times & 50 hours & 100 hours \\
GPUs & 1080Ti*1 & 1080Ti*1\\
Parameters & 55.3M & 262.5M \\
\hline

\end{tabular}
\\
\caption{\label{tab:configuration}
Model configurations
}
\end{table}

\section{Complete Data of Different Domains}
\label{sec:data_domain}

We incorporate different amount of out-of-domain data into in-domain artificial triplets. 
In our experiment, artificial data from IT domain are in-domain when dealing with WMT'18 SMT, while data from Subtitles domain are in-domain when working on SubEdits. 
Complete data on SubEdits and WMT'18 SMT are shown respectively in \ref{tab:detail_domain_subedits} and \ref{tab:detail_domain_smt}.

\begin{minipage} [h]{\textwidth}
\centering
\begin{tabular}{|c|c|c|c|c|} 
\hline
 Out-of-Domain Data & Data Amount &  BLEU↑ & ChrF↑ & TER↓\\
\hline
\multicolumn{2}{|c|}{Only In-Domain Data} & 64.1 & 74.1 & 24.0 \\
\hline
\multirowcell{4}{Medic} & 50K &63.2 &73.3 &24.2 \\
\cline{2-5}
& 100K &63.6 &73.7 &23.9 \\
\cline{2-5}
& 150K &63.7 &73.9 &23.8 \\
\cline{2-5}
& 200K & 63.9& 73.9&23.7 \\
\hline
\multirowcell{4}{Legal} & 50K & 64.2&73.8 &23.4 \\
\cline{2-5}
& 100K & 64.5&73.7 &23.4 \\
\cline{2-5}
& 150K &64.5 &73.9 &23.3 \\
\cline{2-5}
& 200K &\textbf{64.7} &74.3 & \textbf{23.1} \\
\hline
\multirowcell{4}{News} & 50K &63.9 &74.1 &23.6 \\
\cline{2-5}
& 100K &63.6 &73.7 &23.8 \\
\cline{2-5}
& 150K &63.9 &74.1 &23.7 \\
\cline{2-5}
& 200K & 64.2 & \textbf{74.4} & 23.5\\
\hline
\multirowcell{4}{IT} & 50K & 63.2& 73.5&24.2 \\
\cline{2-5}
& 100K &63.6 &73.7 &23.9 \\
\cline{2-5}
& 150K &63.6 &73.6 &24.1 \\
\cline{2-5}
& 200K &63.5 &73.6 &24.0 \\
\hline
\end{tabular}
\\
\captionof{table}{
The experimental results of incorporating different amount of out-of-domain data into in-domain data on SubEdits.
}
\label{tab:detail_domain_subedits}
\end{minipage}

\clearpage
\begin{minipage} [h]{\textwidth}
\centering
\begin{tabular}{|c|c|c|c|c|} 
\hline
 Out-of-Domain Data & Data Amount &  BLEU↑ & ChrF↑ & TER↓\\
\hline
\multicolumn{2}{|c|}{Only In-domain Data} & 68.9 & 84.6 & 19.7 \\
\hline
\multirowcell{4}{Medic} & 50K &67.8 &83.3 &20.9 \\
\cline{2-5}
& 100K &67.9 &83.4 &20.9 \\
\cline{2-5}
& 150K &68.1 &83.6 &20.7 \\
\cline{2-5}
& 200K & 68.3& 83.8&20.0 \\
\hline
\multirowcell{4}{Legal} & 50K & 68.7&83.8 &20.1 \\
\cline{2-5}
& 100K & 69.5&84.4 &19.7 \\
\cline{2-5}
& 150K &69.7 &84.5 &19.5 \\
\cline{2-5}
& 200K &69.7 &84.5 &19.5 \\
\hline
\multirowcell{4}{News} & 50K & 69.7&84.5 &19.5 \\
\cline{2-5}
& 100K &69.7 &84.5 &19.4 \\
\cline{2-5}
& 150K &69.6 &84.4 &19.4 \\
\cline{2-5}
& 200K & \textbf{70.5}& \textbf{84.9} & \textbf{18.9} \\
\hline
\multirowcell{4}{Subtitles} & 50K & 69.5& 84.3&19.4 \\
\cline{2-5}
& 100K &69.4 &84.2 &19.7 \\
\cline{2-5}
& 150K &70.1 &84.6 &19.1 \\
\cline{2-5}
& 200K &69.5 &84.3 &19.7 \\
\hline
\end{tabular}
\\
\captionof{table}{
The experimental results of incorporating different amount of out-of-domain data into in-domain data on WMT'18 SMT.
}
\label{tab:detail_domain_smt}
\end{minipage}

\section{Statistics of Datasets}
\label{sec:statistics}
We split SubEdits, WMT'18 SMT and MLQE-PE into several groups according to the length of \textit{src} and sentence BLEU of \textit{mt}. 
To analyze the performance of our APE model on different groups, we calculate the corpus BLEU score as a measurement. 
The positive value of APE improvement indicates that APE system improves translation quality, whereas APE causes translation quality decline.

\vspace{1cm}
\begin{minipage} [h]{\textwidth}
\centering
\begin{tabular}{|c|c|c|c|c|} 
\hline
Length of \textit{src} & Num & Before APE & After APE & APE improvement \\
\hline
0-10 & 392 & 65.81 & 78.15 & +12.34 \\
\hline
11-20 & 1066 & 62.82 & 73.86 &  +11.04 \\
\hline
21-30 & 498 & 63.36 & 70.86 & +7.50 \\
\hline
>30 & 44 & 64.19 & 70.59 & +6.40 \\
\hline
\end{tabular}
\\
\captionof{table}{
APE performance on different lengths of \textit{src} on WMT'18 SMT.
}
\label{tab:smt_len}
\end{minipage}

\vspace{0.2cm}
\begin{minipage} [h]{\textwidth}
\centering
\begin{tabular}{|c|c|c|c|c|} 
\hline
Length of \textit{src} & Num & Before APE & After APE & APE improvement \\
\hline
0-10 & 4957 & 67.48 & 70.46 & +2.98 \\
\hline
11-20 & 4412 & 60.24 & 63.96 &  +3.72 \\
\hline
21-30 & 591 & 57.93 & 58.99  & +1.06 \\
\hline
>30 & 40 &  44.09 & 45.59 & +1.50 \\
\hline
\end{tabular}
\\
\captionof{table}{
APE performance on different lengths of \textit{src} on SubEdits.
}
\label{tab:subedits_len}
\end{minipage}

\clearpage
\vspace{0.2cm}
\begin{minipage} [h]{\textwidth}
\centering
\begin{tabular}{|c|c|c|c|c|} 
\hline
Length of \textit{src} & Num & Before APE & After APE & APE improvement \\
\hline
0-10 & 124 & 72.17& 72.21 & +0.04 \\
\hline
11-20 & 624 & 73.17 & 72.75 & -0.42\\
\hline
21-30 & 225 & 72.01 & 70.48 & -1.53 \\
\hline
>30 & 7 & 63.57 & 62.32 & -1.25 \\
\hline
\end{tabular}
\\
\captionof{table}{
APE performance on different lengths of \textit{src} on MLQE-PE.
}
\label{tab:mlqe_len}
\end{minipage}

\vspace{1cm}
\begin{minipage} [h]{\textwidth}
\centering
\begin{tabular}{|c|c|c|c|c|} 
\hline
Sentence BLEU Score of \textit{mt} & Num & Before APE & After APE & APE improvement \\
\hline
0-20 & 154 & 11.26 & 34.11 & +22.85 \\
\hline
21-40 & 334 & 33.52 & 51.01 &  +17.49 \\
\hline
41-60 & 474 & 52.61 & 66.98 & +14.37 \\
\hline
61-80 & 512 & 72.00 & 79.25 & +7.25 \\
\hline
81-100 & 526 & 92.64 & 91.64 & -1.00 \\
\hline
\end{tabular}
\\
\captionof{table}{
APE performance on different quality of \textit{mt} on WMT'18 SMT.
}
\label{tab:smt_quality}
\end{minipage}

\vspace{0.2cm}
\begin{minipage} [h]{\textwidth}
\centering
\begin{tabular}{|c|c|c|c|c|} 
\hline
Sentence BLEU Score of \textit{mt} & Num & Before APE & After APE & APE improvement \\
\hline
0-20 & 2120 & 7.37 & 57.47 & +50.10 \\
\hline
21-40 & 1527 & 31.52 & 77.57 &  +46.05 \\
\hline
41-60 & 1201 & 51.86 & 83.87 & +32.01 \\
\hline
61-80 & 650 & 70.48 & 88.82 & +18.34 \\
\hline
81-100 & 4502 & 99.21 & 95.69 & -3.52 \\
\hline
\end{tabular}
\\
\captionof{table}{
APE performance on different quality of \textit{mt} on SubEdits.
}
\label{tab:subedits_quality}
\end{minipage}

\vspace{0.2cm}
\begin{minipage} [h]{\textwidth}
\centering
\begin{tabular}{|c|c|c|c|c|} 
\hline
Sentence BLEU Score of \textit{mt} & Num & Before APE & After APE & APE improvement \\
\hline
0-20 & 62 & 10.31 & 18.84 & +8.53 \\
\hline
21-40 & 114 & 31.48 & 36.50 &  +5.02 \\
\hline
41-60 & 159 & 52.28 & 53.43 & +1.15 \\
\hline
61-80 & 190 & 71.29 & 69.18 & -2.11 \\
\hline
81-100 & 475 & 96.18 & 91.15 & -5.03 \\
\hline
\end{tabular}
\\
\captionof{table}{
APE performance on different quality of \textit{mt} on MLQE-PE.
}
\label{tab:mlqe_quality}
\end{minipage}

\end{document}